\documentclass{article}

\usepackage{microtype}
\usepackage{graphicx}
\usepackage{subfigure}
\usepackage{booktabs} % for professional tables

\usepackage{hyperref}

\usepackage[accepted]{icml2021}

\icmltitlerunning{Curriculum generation using Autoencoder based continuous optimization 
}

\begin{document}

\twocolumn[
\icmltitle{Curriculum generation using Autoencoder based continuous optimization }



\icmlsetsymbol{equal}{*}

\begin{icmlauthorlist}
\icmlauthor{Dipankar Sarkar}{to}
\icmlauthor{Mukur Gupta}{goo}
\end{icmlauthorlist}

\icmlaffiliation{to}{Hike Ltd}
\icmlaffiliation{goo}{None}

\icmlcorrespondingauthor{Dipankar Sarkar}{dipankars@hike.in}
\icmlcorrespondingauthor{Mukur Gupta}{mukur.gupta1@gmail.com}

\icmlkeywords{Curriculum Learning, Deep Learning}

\vskip 0.3in
]



\printAffiliationsAndNotice{} % otherwise use the standard text.

\begin{abstract}
Research in Curriculum Learning have shown better performance on the task by optimizing the sequence of the training data. Recent works have focused on using complex reinforcement learning techniques to find the optimal data ordering strategy to maximize learning for a given network. In this paper, we present a simple yet efficient technique based on continuous optimization trained with auto-encodig procedure. We call this new approach Training Sequence Optimization (TSO). With a usual encoder-decoder setup we try to learn the latent space continuous representation of the training strategy and a predictor network is used on the continuous representation to predict the accuracy of the strategy on the fixed network architecture. The performance predictor and encoder enable us to perform gradient-based optimization by gradually moving towards the latent space representation of training data ordering with potentially better accuracy. We show an empirical gain of 2AP with our generated optimal curriculum strategy over the random strategy using the CIFAR-100 and CIFAR-10 datasets and have better boosts than the existing state-of-the-art CL algorithms.
\end{abstract}

\section{Introduction}

We observe that humans generally learn through a progression of concepts which build on top of each other. This can be viewed in terms level of complexity at each step. We can see this at work right from a baby learning to walk and even take courses online. In both instances, we start with smaller ideas and work towards more complex concepts. In traditional Machine Learning, all the training examples are randomly presented, ignoring the various complexities of the current model's dataset and state. Therefore, it is pertinent to ask if we can use the same learning strategy as humans to improve model training. According to early works \cite{bengio2009} \cite{kumar2010} \cite{Zaremba2014} and recent efforts \cite{fan2018} \cite{graves2017} \cite{guo2020} in various applications of machine learning, it seems this can be the case. This research area is called Curriculum Learning (CL); this can be further decomposed into two aspects. One, where we define the Curriculum, which is the set of tasks a model is trained on while the other is the program where we look at the learning state and choose the model's training tasks. 

A more straightforward definition of CL which was first proposed in \cite{bengio2009} can be understood as training from easier examples to difficult ones. This has been used as a general training strategy for a wide scoping of applications including supervised learning tasks within computer vision \cite{Guo2018}, natural language processing \cite{jiang2014} \cite{plantanios2019} and more. The advantages of CL can be seen as improving the model performance while accelerating the training process, which is very beneficial for Machine Learning research. It can also be seen as a step which is independent of original training algorithms.  

We see many different algorithms being proposed which can be classified within the pre-defined CL, where a human defines the curricula prior and automatic CL, which automatically derives it from the dataset and model. We have seen \cite{tay2019} pre-defined CL or hand-designed curricula used to learn to perform mathematical operations with LSTMs \cite{Hochreiter1997}. Further, \cite{wu2016} used CL to train an RL agent to play Doom and used a small curriculum to improve sample efficiency with imitation learning. All these examples are hand-designed programs, which are fundamentally painful to design and ineffective, leading to catastrophic forgetting with learners \cite{parisi2019}. 

We can say that the pre-defined CL algorithms have a human expert who is the teacher, and the student is the machine learning model. We see the development of automatic CL, where we reduce the dependence on the human expert. The approaches are primarily Self-paced learning \cite{lee2011} \cite{kumar2011}, Transfer teacher \cite{weinshall2018} and RL teacher \cite{graves2017} \cite{matiisen2019}. In recent works, \cite{guy2019} build on the idea of using a scoring function (which is essentially a pre-trained network) to score each training example and then sort them based on their complexity, where they perform a series of experiments with different scoring functions. This technique was shown to boost the speed of training the network as well as the performance. RL teacher algorithms are fully automatic with automated training scheduler compared to the others that are semi-automatic and only address the difficult measure aspect of CL. We also see \cite{yulia2016} where Bayesian Optimization is leveraged to learn the best curricula for word representation learning.

In this work, we take a different view of the CL formulation. We restate it as an optimal training data point problem, for a known task and then address it as an optimization problem. Our approach does not make the easy to hard hypothesis when it comes to curricula generation. We fuse the training scheduler and difficult measure into an embedding that maps to the predicted performance, allowing us to find the optimal sequence. Our contributions are as follows
\begin{itemize}
\itemsep0em 
\item We propose a novel automatic CL approach, Training Sequence Optimization (TSO) where an encoder model maps the training data sequence into a continuous representation. A regression model is built on top to approximate the final performance of architecture. The optimized representation is then used to produce a training data sequence that will perform better. We employ gradient-based optimization techniques which are both fast and powerful.
\item We conduct thorough experiments to show performance boosts when the network is trained with our generated strategy. We show a 2AP performance improvement over the random data ordering and better boosts than the state of the art algorithm. We have performed ablation studies taking into account changes of batch sizes, and network architecture which further showcases this approach's robustness. 
\item We further gain deeper insight into the relationship between the generated strategy and the network size. We show that conventional CL assumptions about easy to hard are not necessarily applicable to all types of networks. Further, we see no transfer of strategy between the same family of networks for a given vision task.
\end{itemize}

\begin{figure}[t]
\vskip 0.2in
\begin{center}
\centerline{\includegraphics[width=\columnwidth]{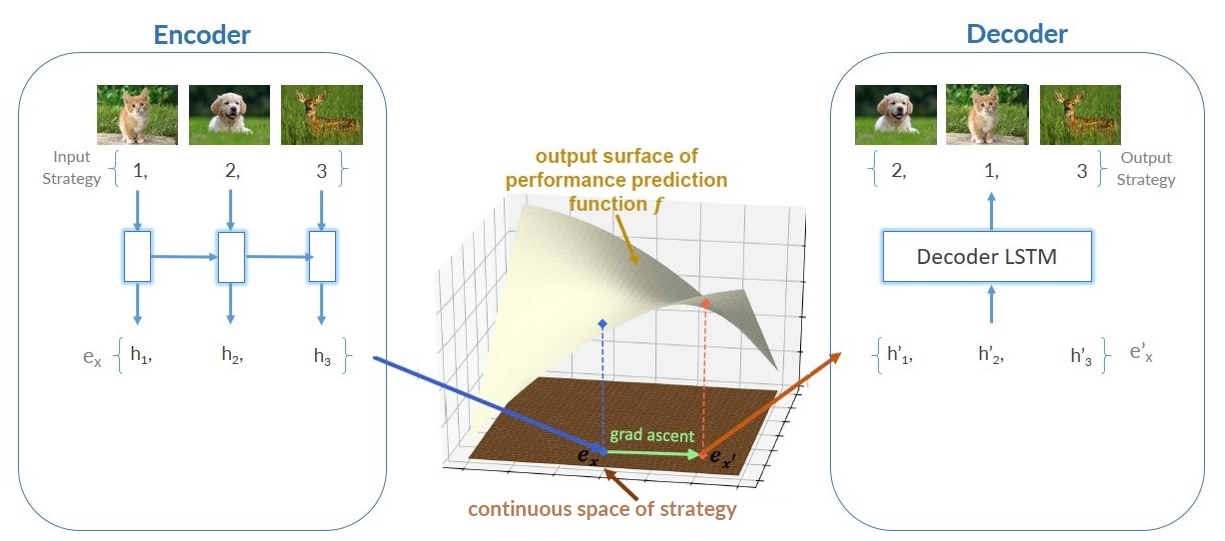}}
\caption{General TSO assembly comprising of an Encoder, performance predictor and a Decoder.}
\label{fig1}
\end{center}
\vskip -0.2in
\end{figure}

\section{Related Works}

Using a curriculum to speed up learning is widely used in human learning and animal training. As the field of deep learning has grown, we see a lot more attention to how data is presented to neural networks while training. An assumption that easy to hard training strategy always helps is seen to be invalidated in research. In \cite{vanya2015}, it is seen that the most learning comes from the hardest examples and neglect easy ones has minimal damage. Further, we see scenarios where the reverse version of CL give better results on specific tasks \cite{zhang2018}. 

We also have hard example mining (HEM) \cite{shrivastava2016}, which is a very well studied area of data selection strategy. Here the assumption is the that hardest examples are the most informative and assigned higher weights. We see the usage of current model losses  \cite{ilya2015} \cite{shrivastava2016} or the gradient magnitude \cite{alain2015} \cite{gopal2016} being used.  We have the boosting algorithm in ensemble learning \cite{yoav1996} that takes the same strategy by uplifting wrongly classified examples.

It is not clear which approach should be applied in what contexts \cite{guy2019}. Recent work has shown that CL works where there are more noisy labels while HEM works for cleaner datasets. We have seen the combination of both strategies as well with a tradeoff policy by the usage of SPL and boosting in \cite{te2016}. However, we can see that fully automatic techniques like RL Teacher would be the better choice between CL and HEM, due to its architecture nature. 

We cannot see CL in isolation, and there is a relation with meta-learning and AutoML where the goal is to find hyperparameters of teaching, including data selection and loss functions. We see the research in Active Learning \cite{burr2009} where a learner aims to achieve higher performance with fewer labelled data by asking for supervision to annotate several unlabelled instances for training. 

\section{TSO: Training Sequence Optimization}

\subsection{Approach}

We propose an iterative gradient-based optimization approach to gradually move towards a better performing training strategy through the iterations. In Figure~\ref{fig1}, there are three major components of TSO. First is the encoder, an LSTM layer which maps a discrete training strategy into its continuous representation. The second component is a performance predictor, essentially a simple feed-forward layer that maps the embedding onto a scalar value, its predicted performance.

We employ gradient-based optimization techniques which are both fast and powerful, to move the embedding in the direction of the increasing gradient of the predicted performance. Among so many strategies possible, predicted performance acts as a heuristics to move in the direction of a strategy expected to perform better. 

To give out the final sequence of training points, a decoder which is also an LSTM layer is used to translate the continuous embedding representation back into the discrete space which is expected to perform better than the original training strategy. Then this TSO assembly is trained, where the performance predictor learns to give a better performance estimate and the decoder learns to map the continuous embedding back into the discrete space. We can break down the complete methodology in 3 parts.

\begin{enumerate}
\itemsep0em 
\item Defining the Curriculum Strategy Space 
\item Different components of TSO
\item Complete algorithm pipeline
\end{enumerate}

\subsection{Defining Curriculum Strategy Space}

We define the search space of the curriculum strategy, which from now on, will be denoted by $X$. In this instance, we consider a computer vision task to demonstrate the boost in the performance of a CNN by the proposed algorithm and to make a fair comparison of the generated strategies, $X$ is initialized with $M$ randomly sampled strategies.

\begin{figure}[t]
\vskip 0.2in
\begin{center}
\centerline{\includegraphics[width=\columnwidth]{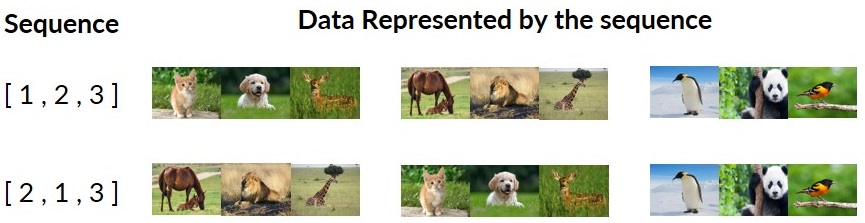}}
\caption{Representation of the training strategy. Several images are batched to make TSO more efficient. Hence we focus on relative ordering of the batches, internal order of the images in a batch is not changed.}
\label{fig2}
\end{center}
\vskip -0.2in
\end{figure}

\textbf{Mapping the image dataset to $X$}: A strategy is a sequence of natural numbers to describe an image dataset where each strategy in $X$ corresponds to a unique relative ordering of images from the data. Therefore each number represents an image, and the sequence represents the relative ordering of the images. So if there are total N images in the training dataset, then $X's$ upper bound would be $factorial(N)$. An example of this shown in Figure~\ref{fig2}. However, such mapping would result in huge sequences and colossal search space which would be computationally difficult to search and optimize. Thus the training dataset is batched to shorten the sequence length and search space, where each number in the sequence represents a group of images, and the order of images within a group is kept constant. So now the problem boils down to optimizing the relative order of the batches. This can be understood from the Figure~\ref{fig2}. We consider $T$ as the final sequence length, where

\begin{equation}
T = N/batch\textunderscore size
\end{equation}

Now we consider a basic CNN and train it on $M$ different strategies from $X$, and then record the validation accuracies of the $M$ trained CNNs corresponding to M different orderings of the training data where the ordering of the validation data is kept the same throughout. So now each strategy $S_{1}$ in $X$ has a performance $Y_{1}$ which is the CNN's validation accuracy trained on $S_{1}$.

\subsection{Components of TSO}

There are three main components in the overall framework(also shown in Figure~\ref{fig1}):
\begin{itemize}
\itemsep0em 
\item Encoder: Maps training strategy into the continuous space
\item Performance Predictor: Predicts the accuracy of the continuous representation of the strategy
\item Decoder: Maps the continuous representation back to the ordered training dataset
\end{itemize}

\textbf{Encoder}: The role of the encoder is to take the sequence representing a strategy and map it into a continuous space to create an embedding. This primarily enables us to do the gradient-based optimization of the input strategy. So for any sequence $S \in X$, the embedding created by the encoder can be represented by $e_{S} = E(S)$. Here, for encoder we use a single layer LSTM and the hidden states of LSTM at each timestamp are used as the continuous representation of S. Hence, $e_{S1}   =   \{h_{1} , h_{2} , h_{3} ........  h_{N}\} \in R^{Nxd}$ where $N$ is the length of input strategy and $h_{t} \in R^{d}$ is the LSTM hidden state at $t^{th}$ timestamp.

\textbf{Performance Predictor}: This component is integrated with the encoder which maps the continuous embedding representation $e_{S}$ onto it’s expected performance $f(E(S))$ which is a scalar value. So firstly, $f$ applies mean pooling operation on all the hidden states of $e_{S} $ to obtain $e_{S1}   =   \frac{1}{T} \sum_{t}^{N} h_{t} $  then use a simple feed forward network to map $e_{S1} $ to it’s predicted accuracy $f(e_{S1}) $. Therefore for a strategy $S \in X$, the aim is to make the predicted accuracy as close as possible to the actual accuracy $(Y)$, which is done by minimizing the least square regression loss $(Y_{S}-f(E(S)))^{2}$

\textbf{Decoder}: The decoder maps the continuous representation of $X$ back to the training strategy in $X$, i.e. taking the embedding generated by encoder $E$ as input to obtain the sequence of training data. Similar to the encoder, we use a single layer LSTM model with the initial hidden state as $h_{N} (S) $which is the hidden state of the last timestamp of the encoder. The decoder works in an auto-regressive manner which is, that it uses the prediction from the previous timestamp to generate the following output. Since the decoder maps $e_{S} $  back into the training strategy so we aim to maximize $log P_{D} (s|E(s)) $ for all the strategies.

\subsection{Training and Inference}

\textbf{Loss Function}: All three components (Encoder, Performance Predictor and Decoder) are trained together and with that we minimize the combination of least square regression loss of performance predictor$(L1)$and the reconstruction loss of the decoder$(L2)$:

\begin{equation}
 L = K L_{1} + (1-K) L_{2}
\end{equation}
\begin{equation}
L1= \sum_{s \in X}^{} (Y_{s}-f(E(s)))^{2}
\end{equation}
\begin{equation}
L2= \sum_{s \in X}^{}log P_{D} (s|E(s))
\end{equation}

Here, $X$ is the set of all the strategies $s$, $K \in $ [0,1] is tradeoff parameter and $Y_{s}$  is the predicted accuracy of the continuous representations of $s$.

This loss is used to train on $M$ training sequences, so with this training framework Encoder learns to create better embedding (continuous representation) of a sequence and Decoder learns to predict back a training sequence given an embedding. After the training is optimized to convergence, better strategies are predicted in the inference mode. So starting with a satisfactory strategy $S$,  we generate it’s continuous representation $e_{S} $  and move this embedding along the gradient direction induced by performance predictor $f$ to finally obtain $e'_{S}$  which is expected to have better accuracy. So:

\begin{equation}
e_{S}   =   \{h_{1} , h_{2} , h_{3} ........  h_{N}\}
\end{equation}
\begin{equation}
e'_{S}   =   \{h'_{1} , h'_{2} , h'_{3} ........  h'_{N}\}
\end{equation}
\begin{equation}
h'_{t} = h_{t} + \eta (\partial f / \partial h_{t})
\end{equation}

where, $\eta $  is the step size. Then $e'_{S} $  is fed as input to the decoder to convert it back to the training strategy $S'$ which is probably better than $S$ because $f(E(S`)) \geq f(E(S)) $.The whole process is summed up in the Algorithm~\ref{alg:example}:

\begin{algorithm}[tb]
   \caption{TSO : Training Sequence Optimization}
   \label{alg:example}
\begin{algorithmic}
   \STATE {\bfseries Input:} Initialize $X_{eval}$ with $M$ randomly sampled strategies, $Y_{eval}=\phi$. Training set for TSO $X_{train}=\phi$, $Y_{train}=\phi$ . Step size $\eta $. Number of optimization iterations $L$. Network Architecture $N1$. Encoder $E$, performance predictor $f$ and Decoder $D$.
   \FOR{$i=1,....L$}
   \STATE Train $N1$ on each $s \in X_{eval}$ to obtain it's validation performance $Y_{eval}=\{Y_{s}\}, \forall s\in X_{eval}$
   \STATE Enlarge $X_{train}:X_{train}\cup X_{eval}$
   \STATE Enlarge $Y_{train}:Y_{train}\cup Y_{eval}$
   \STATE Train TSO assembly using $X_{train}$ and $Y_{train}$
   \STATE Pick $K$ strategies with top $K$ performances from $X_{train}$ to form $X_{K}$
   \STATE For $s\in X_{K}$ obtain better representation by moving $e_{s}$ in the gradient direction induced by $f$ to get $e'_{s}$
   \STATE Use decoder to obtain $X'_{K}=\{s',\forall s'=D(e'_{s})\}$
   \STATE Update $X_{eval}=X'_{K}$
   \ENDFOR
   \STATE {\bfseries Output:} Top performing strategies in $X_{train}$
\end{algorithmic}
\end{algorithm}

\section{Experiments}

\begin{table}[t]
\caption{Image datasets used in the experiments.}
\label{dataset-table}
\vskip 0.15in
\begin{center}
\begin{small}
\begin{sc}
\begin{tabular}{lcc}
\toprule
Data set & Classes & Total Samples \\
\midrule
CIFAR-10           &10           &50000\\
Sampled-CIFAR-100  &5            &2500\\
MNIST              &10           &60000\\
\bottomrule
\end{tabular}
\end{sc}
\end{small}
\end{center}
\vskip -0.1in
\end{table}

\begin{table*}[ht]
\caption{Performance boost by TSO over the random data ordering after the $1^{st}$ and $2^{nd}$ iterations on various datasets and  architectures.}
\label{results-table}
\vskip 0.15in
\begin{center}
\begin{small}
\begin{sc}
\begin{tabular}{lccccc}
\toprule
Dataset & Network Architecture & Random Strategy & Iteration1 & Iteration2\\
\midrule
CIFAR-10             &N1     &60.78    &61.00    &61.18\\
CIFAR-10             &N2     &63.37    &63.71    &63.79\\
MNIST                &N3     &96.95    &97.06    &97.07\\
Sampled CIFAR-100    &N4     &47.92    &48.92    &48.97\\
Sampled CIFAR-100    &N5     &50.39    &51.98    &51.93\\
\bottomrule
\end{tabular}
\end{sc}
\end{small}
\end{center}
\vskip -0.1in
\end{table*}

In this section, we highlight the ablation analysis and TSO's empirical performance on a computer vision task in different settings.

\subsection{Datasets}

We perform several experiments on following datasets which are also outlined in Table~\ref{dataset-table}:
\begin{itemize}
\itemsep0em 
\item \textbf{MNIST}: This dataset \cite{lecun1998} is a collection of 70,000 images divided into 60,000 training images and 10,000 testing images across 10 categories. It contains 28x28 pixel grayscale images representing different handwritten digits from 0 to 9, thus having ten different categories.
\item \textbf{CIFAR-10:} This dataset \cite{alex2009} contains 32x32 pixels coloured 60,000 images divided into 50,000 training images, and 10,000 testing images. They are spread equally across ten categories: aeroplane, automobile, bird, cat, deer, dog, frog, horse, ship and truck.
\item \textbf{Sampled-CIFAR-100}: This dataset is similar to CIFAR-10 with 50,000 training images and 10,000 testing images spread equally across 100 classes which are grouped into 20 super-classes. We consider small mammals super-class of CIFAR-100, a subset of 3000 images spread equally across five classes.
\end{itemize}

\subsection{Settings}

We depict the performance boost by TSO on three datasets using 5 CNNs (Table~\ref{cnn-table}): $N1, N2, N3, N4$ and $N5$ and as shown in Table~\ref{testcases-table}, we define 4 test cases to conduct an ablation analysis. The batch size to reduce the strategy length and search space-bound was set at 500 in most of the experiments to get a training strategy of length 100. We start with 150 randomly sampled training sequences in $X$ set and pick top 50 sequences in each iteration from the training set for the inference process to generate potentially better strategies and then merge them into the training set for the next iteration. Therefore, for CIFAR-10 with strategy length 100 will have the following parameters N = 		50000, M = 150, Batch size 	= 500, Validation size = 10000, and Sequence length(T) = (50000/500) = 100.

The architecture of the encoder is a single layer LSTM where sequence length is 100 with embedding size and hidden state size as 32 and 96 respectively. Respective hidden states of the encoder at different time stamps are normalized, which constitute the embedding of a strategy: $e_{S1}   =   \{h_{1} , h_{2} , h_{3} ........  h_{N}\} $. The performance predictor $f$  is a single layer feed-forward network which takes $\frac{1}{T} \sum_{t}^{N} h_{t} $  as input and produces a scalar value as the predicted accuracy. 

Finally, the decoder is also a single layer LSTM where total timestamps are 100 with hidden state size 128. The whole assembly is then trained on Adam optimizer with the following hyper-parameters: Epochs = 1000, Learning Rate = 0.001 and Trade-off(K) = 0.65.

\subsection{Results}

\begin{table}[t]
\caption{Handcrafted CNN Architectures used.}
\label{cnn-table}
\vskip 0.15in
\begin{center}
\begin{small}
\begin{sc}
\begin{tabular}{cccc}
\toprule
Network & Layers & Optimizer &Total Parameters \\
\midrule
N1           &8           &Adam   &933,290\\
N2           &6           &Adam   &923,914\\
N3           &5           &SGD    &36,901\\
N4           &13          &SGD    &159,901\\
N5           &20          &SGD    &1,208,101\\
\bottomrule
\end{tabular}
\end{sc}
\end{small}
\end{center}
\vskip -0.1in
\end{table}

\begin{figure}[t]
\vskip 0.2in
\begin{center}
\centerline{\includegraphics[width=\columnwidth]{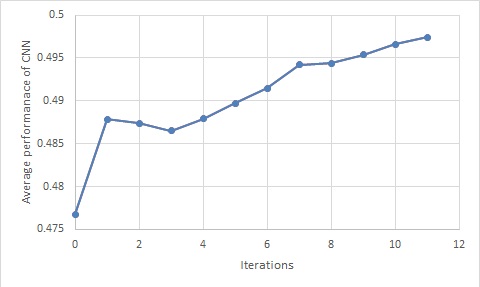}}
\caption{Illustrating the average performance boost of newly generated strategies over the random data ordering (iteration-0) on the Sampled-CIFAR-100 dataset on N3 network architecture.}
\label{fig3}
\end{center}
\vskip -0.2in
\end{figure}

\begin{figure}[t]
\vskip 0.2in
\begin{center}
\centerline{\includegraphics[width=\columnwidth]{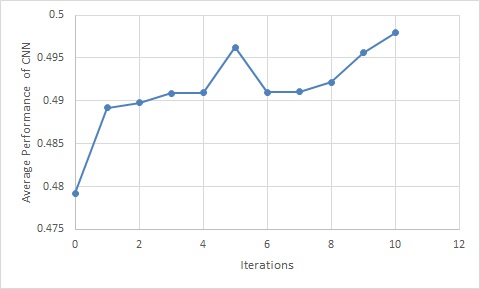}}
\caption{Illustrating the average performance boost of newly generated strategies over the random data ordering (iteration-0) on the Sampled-CIFAR-100 dataset on N4 network architecture.}
\label{fig4}
\end{center}
\vskip -0.2in
\end{figure}

We start with 150 randomly sampled strategies and generate 50 new better performing strategies in subsequent iterations. In Table~\ref{results-table}, we show the average performance of 50 new generated strategies after $1^{st}$ and $2^{nd}$ iterations and observe a consistent performance boost compared to when the network is trained with randomly ordered data. We further conduct two additional experiments up to $10^{th}$ iteration to observe a similar gradient ascent behaviour and consistent performance surge throughout iterations starting with random data ordering. This is shown in figure~\ref{fig3} and figure~\ref{fig4}.

\begin{table}[ht]
\caption{Relative percentage boost comparison of TSO and the approach by \cite{guy2019} on random data ordering. Test case was performed on Sampled CIFAR-100 dataset on N5 network architecture.}
\label{guy-compare}
\vskip 0.15in
\begin{center}
\begin{small}
\begin{sc}
\begin{tabular}{cc}
Approach& Relative Percentage boost\\
\midrule
\citeauthor{guy2019}             &2.70\\
TSO                        &3.45\\
\bottomrule
\end{tabular}
\end{sc}
\end{small}
\end{center}
\vskip -0.1in
\end{table}

Moreover, in Test case-1, we try to replicate the dataset and network architecture used by \cite{guy2019} and demonstrate more relative percentage boost in the performance at the end of $3^{rd}$ iteration. This is also shown in Table-~\ref{guy-compare}.

\begin{table*}[ht]
\caption{To validate the performance and robustness we define following test cases varying the dataset, Network Architecture and the Strategy length.}
\label{testcases-table}
\vskip 0.15in
\begin{center}
\begin{small}
\begin{sc}
\begin{tabular}{cllccc}
\toprule
Test case & Dataset(s) & Network(s) & Strategy Length(s) & Step Size($\lambda$) & Total images(N)\\
\midrule
1  &Sampled CIFAR-100  &N5        &100     &5    &2500\\
2  &CIFAR-10           &N1,N2     &100     &7    &50000\\
3  &CIFAR-10           &N2        &100,200 &7    &50000\\
4  &Sampled CIFAR-100  &N3,N4,N5  &100     &5    &2500\\
\bottomrule
\end{tabular}
\end{sc}
\end{small}
\end{center}
\vskip -0.1in
\end{table*}

\subsection{Ablation Study}

We perform an ablation study with the 4 test cases, as shown in Table~\ref{testcases-table}. We vary the network architecture and change the strategy length to further assess the robustness of our approach.

\subsubsection{Varying Network Architecture}

We demonstrate TSO's performance gains by varying the Network Architecture in the $2^{nd}$ test case. We see an increase in the mean performance of the newly generated strategies over the iterations despite the change in the CNN complexity. 

This behaviour can be seen in Table~\ref{network-ablation-table} with where CIFAR-10 dataset is kept constant (and using Networks N1 and N2) and also in figure-~\ref{fig3} and figure-~\ref{fig4} where Sampled CIFAR-100 is constant, and Networks N3 and N4 are varied.

\begin{table}[h]
\caption{Illustrating the performance boost over the iterations for different network architectures on same dataset.}
\label{network-ablation-table}
\vskip 0.15in
\begin{center}
\begin{small}
\begin{sc}
\begin{tabular}{cccc}
Network & Random Strategy & Iter1 & Iter2\\
\midrule
N1             &60.78    &61.00    &61.18\\
N2             &63.37    &63.71    &63.79\\
\bottomrule
\end{tabular}
\end{sc}
\end{small}
\end{center}
\vskip -0.1in
\end{table}

\subsubsection{Varying Strategy Length}

\begin{table*}[h]
\caption{Illustrating the effect  of strategy length (or batch size) on the performance boost. Performance grows a lot faster in case of longer strategy but also doubles the time taken.}
\label{strategy-len-table}
\vskip 0.15in
\begin{center}
\begin{small}
\begin{sc}
%\begin{tabular}{p{0.2\linewidth}p{0.15\linewidth}p{0.15\linewidth}p{0.15\linewidth}p{0.1\linewidth}p{0.1\linewidth}p{0.1\linewidth}}
\begin{tabular}{ccccccc}
Strategy Length & Batch Size & Time per step(mins) & Initial performance &Iter1 & Iter2 &Iter3\\
\midrule
200             &250     &42    &63.58    &63.74   &63.86   &63.87\\
100             &500     &23    &63.60    &63.64   &63.78   &63.83\\
\bottomrule
\end{tabular}
\end{sc}
\end{small}
\end{center}
\vskip -0.1in
\end{table*}

In the $3^{rd}$ test case, we vary the batch size, which we used to shorten the sequence length and make the algorithm more efficient. Change in batch size leads to change in final sequence length, and we observed an apparent increase in performance as the strategy length was increased, which is following the fact that the longer strategy can hold more information. This is shown in Table~\ref{strategy-len-table}, where we halved the batch size to double the sequence length to 200 and tested the algorithm on the CIFAR-10 dataset. Although the strategies with larger batch size were at a higher performance at initialization, the performance of lower batch size strategies grew a lot faster and ended with better performance after four iterations. However, this also almost doubles the time taken to complete a step.

\section{Observations}
\subsection{Performance boost}

In Table~\ref{results-table}, we demonstrate that the mean performance of 50 new generated sequences after each iteration keeps increasing. This increase in performance over the iterations is shown on three different datasets and five different CNNs. This increasing performance is also portrayed in figure~\ref{fig3}, figure~\ref{fig4} and Table~\ref{strategy-len-table} with different settings.

The proposed algorithm takes on an average less than 0.5 GPU hour to complete each iteration which is a lot more efficient than all the previous attempts at Curriculum Learning involving deep neural networks. In comparison, \cite{guy2019} employ a scoring function and a pacing function to score each training example on difficulty level and then sample mini-batches that show an increasing level of difficulty, which is very inefficient and time-consuming.

\subsection{Strategy transfer performance}

The space of Curriculum Learning has been quite widely explored using varying approaches like by \cite{weinshall2018}, and \cite{guy2019}, but there have been no comments around the Curriculum’s dependence on the Network architecture. We believe that the difficulty cannot be absolute and is a perception of the learner, and thus the Curriculum cannot be the same for different learners. 

In our work, we empirically show that the optimal order of the training examples depends not just on the training examples but also on the learner, i.e. the network architecture. To show this, we calculate the average distance between the strategies in the final outputs produced by two different networks. We calculate the distance between two strategies S1 and S2 using:

\begin{equation}
    \sum_{i=1}^{100} |S_{1}(i)-S_{2}(i)|
\end{equation}

where, $S_{1}(i)$ represents position of i in $S_{1}$.
The mean distance between networks N1 and N2 on CIFAR-10 was 32.8 which shows that the optimal strategies generated by both the networks are quite different, as, in case of a sequence of length 100, maximum and minimum possible distances are 0 and 50.5 respectively. This observation is following the fact that as the layers were added to the network, it becomes more capable of holding more information and thus becomes more ignorant to the training dataset's ordering.

Further, we transfer the discovered optimal strategies of N1 to evaluate them on N2 and vice-versa. We see a deficient performance in both cases. This again proves network dependency of the discovered curriculum strategies.

\subsection{Network complexity}

In test case 4, we explore the relationship between the increase in accuracy by TSO and network complexity. From Table~\ref{cnn-table}, the complexity of networks vary in following fashion: $N5>N4>N3$. In the following Table~\ref{network-var-table}, we can see that the surge in performance reduces as the network grows in size. This observation agrees with our earlier hypothesis that the optimal training strategy or a curriculum depends on the learning network. Moreover, as the learner grows in size, it becomes capable of holding more information and hence starts becoming ignorant to the data order.

\begin{table}[h]
\caption{Variation of performance boost with network architecture complexity. This test case was performed on Sampled  CIFAR-100 dataset.}
\label{network-var-table}
\vskip 0.15in
\begin{center}
\begin{small}
\begin{sc}
\begin{tabular}{cc}
Network Architecture& Performance boost in A.P.\\
\midrule
N3             &2.07\\
N4             &1.8\\
N5             &1.74\\
\bottomrule
\end{tabular}
\end{sc}
\end{small}
\end{center}
\vskip -0.1in
\end{table}

\section{Conclusion}

We employ a gradient-based optimization to find a strategy viz Curriculum in which the training examples should be presented to the learner to ensure better performance and faster convergence. We propose an iterative algorithm that performs a lot faster, boosting the performance after each iteration. 

With pre-defined CL and even, automatic CL, we see the assumption of the intuitive understanding of the Curriculum, based on difficulty. In \cite{guy2019}, they sort the data points based on a difficulty score and then sample mini-batches representing increasing order of difficulty. Their work is as per the intuition behind the Curriculum but has no mathematical basis, i.e. why do the training examples sorted by the scoring function perform better than the random order. Our findings indicate that we cannot transfer strategy even for simple network architecture families for the same task, thus indicating the theoretical limitations of existing approaches modelling it as a transfer learning problem.  

Our algorithm does not make any assumptions on the Curriculum or difficulty. It optimizes the strategy's performance using the gradient ascent technique and moves towards optimality with each iteration. We further establish that the Curriculum for a dataset is not absolute but is dependent on the learner network. Our findings are that smaller networks are typically more sensitive to the strategy, which would make it useful for improving the performance of resource-constraint models without any architecture changes. 

We see our approach being applicable across other tasks beyond vision and expect to see improvement gains elsewhere. We would want to explore other optimization techniques and assess their performance as well.


\bibliography{main}
\bibliographystyle{icml2021}

\end{document}